\documentclass[runningheads]{llncs}

\usepackage{eccv}

\usepackage{eccvabbrv}

\usepackage{graphicx}
\usepackage{booktabs}
\usepackage{svg}
\usepackage{amsmath}
\usepackage{pifont}

\usepackage[accsupp]{axessibility}  

\usepackage[pagebackref,breaklinks,colorlinks,citecolor=eccvblue]{hyperref}
\usepackage{hyperref}

\usepackage{orcidlink}

\newcounter{ablationpara}[section]
\renewcommand{\theablationpara}{\thesection.\arabic{ablationpara}}
\newcommand{\ablationhead}[1]{\refstepcounter{ablationpara}\noindent\textbf{\theablationpara\quad #1:}}

\begin{document}

\title{GoDeep: Annotation-Free Open-Vocabulary 3D Scene Understanding via Language-Space Lifting} 

\titlerunning{GoDeep: Anot. free OVoc 3D Sc Und. via Lang. Space Lifting}

\author{Thodoris Betsas\orcidlink{0000-0001-7354-5544} \and
Anastasios Doulamis\orcidlink{0000-0002-0612-5889} \and
Andreas Georgopoulos\orcidlink{0000-0001-6520-9954}}

\authorrunning{T.~Betsas et al.}

\institute{ Laboratory of Photogrammetry, School of Rural, Surveying and Geoinformatics Engineering, NTUA, 15772 Athens, Greece \\
\email{betsasth@mail.ntua.gr, adoulam@cs.ntua.gr, drag@central.ntua.gr}}

\maketitle

\begin{abstract}
Open vocabulary 3D semantic segmentation methods typically lift CLIP features
into 3D. This embeds points in a joint vision-language space known
to behave like a bag-of-words on compositional tasks. Furthermore, even
annotation free variants often require a large 3D training corpus
and a dedicated 3D encoder per domain. Instead we use a
vision-language model purely as a translator. It produces
structured, entity-level descriptions of each posed image. These
descriptions are grounded, projected, and aggregated directly in a
general-purpose, language-only embedding space, with no 3D training
corpus or encoder required. On ScanNet++, our pipeline is competitive with strong
annotation free baselines trained on ScanNet. On a 5-building
cultural heritage benchmark, raw scores initially favor a CLIP-based
variant, but a single systematic vocabulary correction reverses this
ranking. An effect confirmed by a second, independent correction on
a different class, indicating that language-space embeddings track
physical content more faithfully. This fidelity extends to genuinely out-of-vocabulary (OOV) objects on
ScanNet++ proving that language-space embeddings separate presence from absence objects far more sharply than CLIP-based embeddings do. GoDeep also
localize these OOV objects within the scene, all without any 2D-3D annotation. Because every
representation remains discrete text, predictions are also
explainable at the point level. Finally, exploiting both a heuristic weighting, that favors precise over merely frequent observations and GoDeep's explainability property, we propose an aggregation strategy, as a proof of concept, that favors finer elements localization.

\end{abstract}

\section{Introduction}
\label{sec:intro}
3D semantic segmentation has traditionally relied on human-annotated
point clouds, a process that is costly and labor-intensive to carry
out at the point level, and that, even when done carefully, rarely
captures every object category actually present in a scene. As a
result, the classes used by mature closed-set 3D segmentation methods
\cite{wu2024point, wu2022point, milioto2019rangenet++} are typically
coarse structural categories like wall, column and roof, into which
finer elements (a capital, a shaft, a frieze) are silently absorbed,
losing their own semantic identity \cite{matrone2020benchmark, betsas2026exploring}. The
dominant recipe for open-vocabulary 3D segmentation addresses this by
lifting 2D vision-language features into 3D, either by distilling CLIP
\cite{radford2021learning} embeddings onto point clouds
\cite{peng2023openscene} or by training a 3D encoder against 2D or
textual pseudo-supervision \cite{yang2024regionplc, Lee_2025_CVPR,
Li_2025_ICCV}. These methods share an assumption rarely questioned: that 3D
points should be embedded in CLIP's joint vision-language space, a
space shown to behave like a bag-of-words on relational and
attribute-binding tasks \cite{yuksekgonul2023bags, Thrush_2022_CVPR},
a liability precisely where open-vocabulary segmentation matters
most, e.g., in cultural heritage documentation.

Despite avoiding human-annotated 3D labels, these methods remain
resource-dependent in a different sense: each still requires
assembling a large, purpose-built 3D training corpus, paired with
dense 2D vision-language supervision \cite{peng2023openscene,
yang2024regionplc, Lee_2025_CVPR, Li_2025_ICCV}, to train or distill
a dedicated 3D encoder for every target domain, and queries are then
mediated by the same limited CLIP text encoder discussed above. At
the same time, modern VLMs are already capable of producing dense,
structured scene descriptions \cite{wang2024qwen2}, yet this richness
is seldomly preserved in the resulting 3D representation, typically
reduced to whatever a short class name or generic caption can convey
— a limitation further compounded by CLIP's 77-token context window.

Rather than lifting CLIP features, we use a vision-language model
(e.g., Qwen2-VL \cite{wang2024qwen2}) purely as a translator: applied
independently to each posed image of a scene, it produces a
structured, entity-level description of that view. Each described
entity is grounded back into the image via an open-vocabulary
segmenter (SAM3 \cite{carion2025sam}) and projected onto the point
cloud using the camera's known pose, so that every 3D point
accumulates the set of natural-language entities observed for it
across views. These per-point entity sets are then aggregated
directly in a general-purpose sentence embedding space
\cite{reimers2019sentence}, without a limitation to the number of entities or words. The resulting per-point features
therefore live entirely in this language-only space, rather than any
joint vision-language space: alignment performed internally by the
2D segmenter is confined to grounding text onto pixels and never
propagates into the 3D representation itself. The pipeline needs no
3D training corpus and no dedicated 3D encoder, since every component
is a frozen, off-the-shelf 2D or language model, applied directly to
a new scene at inference time.

We evaluate this design on two fronts. On the 100-class ScanNet++
benchmark~\cite{Yeshwanth_2023_ICCV}, our pipeline is competitive
with strong annotation-free baselines trained on ScanNet, though it
falls behind to methods trained on larger multi-dataset corpora or
directly on the target distribution (Section~\ref{subsec:results}). On a multi-building
cultural heritage dataset~\cite{pellis2025photogrammetric}, raw benchmark scores initially favor the CLIP-based variant.
Paragraph~\ref{para:semantic-truth} shows that correcting a single systematic vocabulary
mismatch, dataset-wide, improves the language-space representation
far more than its CLIP-based counterpart, reversing the ranking. Our contributions are fourfold:
\begin{itemize}
    \item We present an annotation-free 3D scene understanding
    pipeline that decouples 3D representation learning from CLIP's
    joint embedding space by lifting structured, VLM-generated
    descriptions into a pure sentence-embedding space.
    \item We show this design matches 3D-trained annotation-free
    baselines on a standard indoor benchmark without any 3D training.
    \item We show, on a real 5-building cultural heritage benchmark, that correcting a single systematic vocabulary mismatch reverses the ranking between language-space and CLIP-based embeddings, evidence that the former tracks physical content more faithfully, a property that becomes a liability only when the evaluation vocabulary itself is imprecise.
    \item We show the pipeline is explainable at the per-point level: predictions trace back to specific, weighted natural-language evidence, and this weighting measurably favors precise observations over merely frequent ones.
\end{itemize}

\section{Related Work}
\label{sec:relwork}

Closed-set 3D semantic segmentation algorithms can be classified by
representation into point-based \cite{zhao2021point, wu2022point,
qi2017pointnet++}, dimensionality-reduction \cite{li2021multi,
milioto2019rangenet++, xiao2021fps}, discretization-based
\cite{choy20194d}, graph-based \cite{wang2019dynamic}, and hybrid
\cite{wu2024point, ye2023uniseg, yue2025litept} methods
\cite{betsas2025deep}. These methods achieve strong results but are
confined to the fixed class vocabulary seen during training, and are
further sensitive to the acquisition modality of the training data
(e.g., RGB-D, LiDAR etc.), limiting their transferability across
sensor types and domains \cite{betsas2025deep}.

Inspired by 2D open-vocabulary segmentation
\cite{hu2016segmentation, ghiasi2022scaling, li2022language}, 3D
open-set scene understanding methods utilize vision-language models
to move beyond closed class sets. OpenScene \cite{peng2023openscene}
and PLA \cite{ding2023pla} distill CLIP features or hierarchical
3D-caption pairs onto point clouds, training a dedicated 3D encoder
in each case. OpenMask3D \cite{takmaz2023openmask3d} instead pairs a
once-trained, class-agnostic 3D instance mask proposal network with
CLIP embeddings computed at inference. CLIP-FO3D
\cite{Zhang_2023_ICCV} similarly distills dense CLIP features into a
trained 3D encoder. More recently, Mosaic3D \cite{Lee_2025_CVPR} and
SceneSplat \cite{Li_2025_ICCV} scale this recipe with larger
VLM-generated pseudo-label corpora, achieving strong results on
ScanNet++, while RegionPLC \cite{yang2024regionplc} does so on
ScanNet and ScanNet200. All of these methods query through CLIP's
joint vision-language embedding space and, except for OpenMask3D's
frozen mask backbone, require training a dedicated 3D encoder per
target domain, a design we revisit in Section~\ref{sec:method}.

This shared reliance on CLIP's text encoder is not incidental: CLIP
is trained with a contrastive image-text objective in which a short
caption need only distinguish its paired image from others in a
batch, a task solvable by recognizing salient keywords without
resolving how they relate to one another. Sentence encoders such as
the one we use \cite{reimers2019sentence} are instead trained on
natural language inference and semantic textual similarity, tasks
that directly reward correctly parsing relations between words. This
distinction is measurable: CLIP-family encoders behave like a bag-of-words on tasks requiring attribute
binding and relational reasoning \cite{yuksekgonul2023bags,
Thrush_2022_CVPR}. We revisit this distinction empirically in
Section~\ref{para:semantic-truth}.

\begin{figure}[t]
  \centering
  \includegraphics[width=1\textwidth]{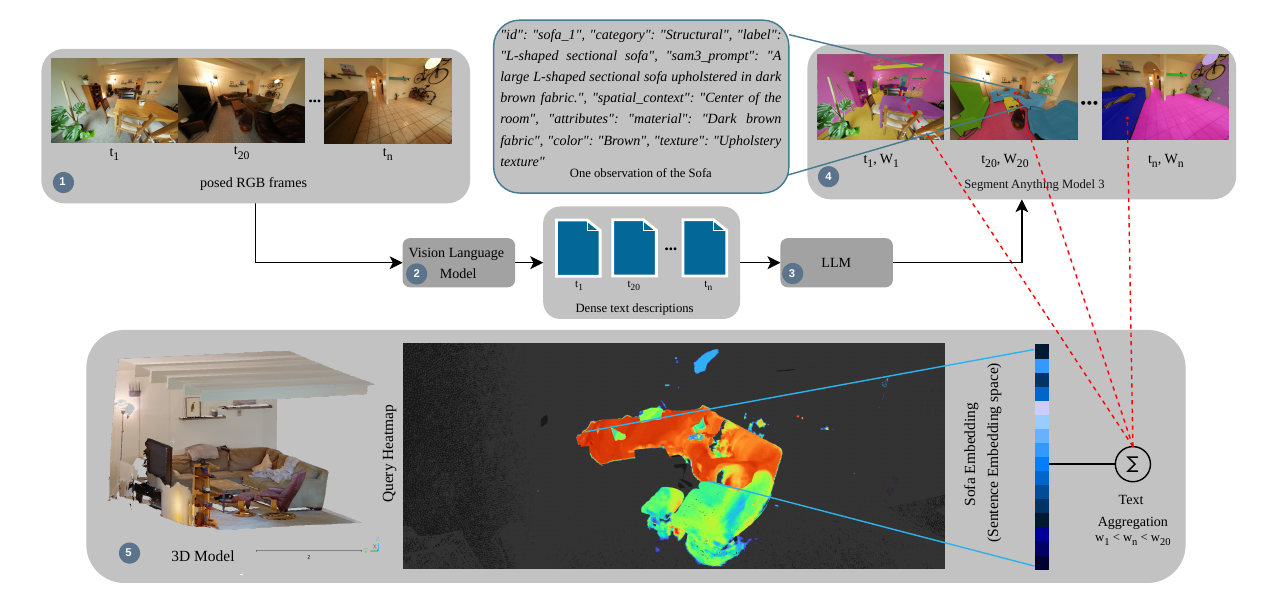}
  \caption{Overview of the GoDeep pipeline on a single query
(\emph{sofa}): (1) posed frames are (2) described by a
vision-language model and (3) structured by an LLM into JSON
entities. (4) Each entity's prompt grounds a mask via
SAM3~\cite{carion2025sam}, weighted by depth, centrality, and mask
precision ($w_1, w_2, w_3$). (5) Weighted observations are combined
via mean-inliers pooling into a per-point sentence embedding, stored
on the point cloud and compared against a text query to produce the
heatmap shown (blue: low, red: high similarity).}
  \label{fig:GoDeep_Pipeline}
\end{figure}

\section{Method}
\label{sec:method}

\subsection{Overview}
\label{subsec:overview}
Figure~\ref{fig:GoDeep_Pipeline} summarizes our pipeline. A
vision-language model describes each posed image of a scene in
natural language. An open-vocabulary segmenter grounds each described
entity to pixels. Each 3D point is then projected into its nearby
camera views to collect the entity groundings that cover it. The
resulting per-point natural-language evidence is aggregated in a
sentence embedding space. No stage is trained on the target scene or
domain. Every intermediate representation, including the VLM
description, the JSON entity, and the aggregated embedding, also
remains a discrete, human-readable piece of text. This lets us trace
any point's final label back to the specific sentences that produced
it, a property we exploit directly in Paragraph~\ref{para:explainability}.

\subsection{Structured Description Generation and Grounding}
\label{subsec:structdesc}
For each posed image, a vision-language model (Sec.~\ref{sec:intro})
produces a free-form dense description of the visible content. A
lightweight language model then converts this description into a
structured JSON list of entities. This step filters out content
irrelevant to the task at hand, e.g., sky or lighting conditions in an
architectural survey. Each entity retains a category, a short label, a
segmentation-oriented prompt, and free-text attributes (material,
color, texture) when mentioned (Figure~\ref{fig:GoDeep_Pipeline}).
Entities with associated defects or conditions are recorded
separately. Each entity's prompt is then passed to
SAM3~\cite{carion2025sam}. Masks are not mutually exclusive: a point
may accumulate evidence from multiple overlapping entities, e.g., a
structural element and a defect on its surface.

\subsection{Multi-View 3D Lifting and Language-Space Aggregation}
\label{subsec:mv3dlift}
Each 3D point is projected into its candidate camera views using the
camera's calibrated intrinsics, pose, and lens distortion model
(radial-tangential, following the standard OpenCV parameterization).
For each camera $c$, we build a downsampled depth buffer $Z_c$ from
the point cloud itself, and discard a point $p$ as occluded from $c$
if its camera-space depth $z$ exceeds the buffered depth at its
projected location $\pi(p)$ by more than a fixed tolerance $\tau$.
Among all candidate cameras, we first keep only the $K$ closest to
$p$ by raw depth, then discard any of these $K$ that fail the
occlusion test above. Therefore, a point may receive descriptions from
fewer than $K$ cameras when some of its nearest views are occluded.

Each visible (point, camera, mask) triple contributes a weight
\begin{equation}
\label{eq:weight}
w \;=\; \frac{1}{z+\epsilon} \; \cdot \; \exp\!\left(-\frac{d_{2D}^2}{d_{\max}^2}\right) \; \cdot \; \frac{1}{\sqrt{a}+\epsilon}.
\end{equation}
Here, $z$ is the depth in camera $c$, $d_{2D}$ is the 2D distance to
the image center, $d_{\max}$ is a normalizing constant, $a$ is the
mask's relative area, and $\epsilon$ is a small constant for numerical
stability. The three factors favor closer, more centrally-imaged, and
more precisely-segmented observations. Peripheral, distant, or
coarsely segmented masks are treated as less reliable evidence.
Paragraph~\ref{para:explainability} shows this weighting is not a minor
implementation detail: it measurably reorders which evidence
dominates a point's final label.

Each point accumulates a weighted set $\{(v_i, w_i)\}$ of
natural-language entity embeddings. Here $v_i \in \mathbb{R}^d$ is the
sentence embedding \cite{reimers2019sentence} of the $i$-th observed
entity, and $w_i$ is its weight from above. We aggregate this set via
mean-inliers pooling: a first-pass weighted mean $\bar{v}$ is
computed, then every observation is compared against this same
$\bar{v}$ via cosine similarity, and the final mean $\hat{v}$ is
recomputed using only the observations that pass this test:
\begin{equation}
\bar{v} = \operatorname{normalize}\Big(\textstyle\sum_i w_i v_i\Big),
\qquad
\hat{v} = \operatorname{normalize}\!\Big(\textstyle\sum_{i \,:\, v_i \cdot \bar{v} > \delta} w_i v_i\Big).
\end{equation}
Observations with similarity below a fixed threshold $\delta$ to
$\bar{v}$ are discarded as outliers, and $\hat{v}$ is recomputed over
the remaining inliers. This scheme absorbs occasional VLM
hallucinations or mis-grounded masks without letting a single bad
observation dominate a point's representation. Section~\ref{par:aggcompar}
compares it against four alternative aggregation strategies. Points
with no covering observation are inpainted from their $k$ nearest
neighbors with a valid embedding, weighted by inverse distance, and
the aggregated embeddings are optionally projected to a lower
dimension via PCA.

\section{Experiments}
\label{sec:experimental_setup_results}

\subsection{Setup}

\textbf{Datasets.} We evaluate on ScanNet++~\cite{Yeshwanth_2023_ICCV}
(50 validation scenes, top-100 class benchmark) and on a cultural
heritage dataset~\cite{pellis2025photogrammetric} spanning 5 historic
buildings captured with a mix of terrestrial laser scanning and
photogrammetry, annotated with the 10 ARCH-standard
classes~\cite{matrone2020benchmark}.

\textbf{Evaluation Protocol.} We assign each point its argmax class by
cosine similarity between the point's aggregated embedding and the text embeddings of the class names, following the protocol used by
Mosaic3D~\cite{Lee_2025_CVPR} and SceneSplat~\cite{Li_2025_ICCV}. No
per-class thresholds are involved. On ScanNet++, we report f-mIoU and
f-mAcc, excluding wall, floor, and ceiling, following standard
practice for this benchmark. On the heritage dataset, we report mIoU,
mAcc, and mPrec over all 10 ARCH classes, including the catch-all
\emph{other} category.

\textbf{Baselines.} On ScanNet++, we report literature numbers for
OpenScene~\cite{peng2023openscene}, RegionPLC~\cite{yang2024regionplc}, Mosaic3D~\cite{Lee_2025_CVPR}, and
SceneSplat~\cite{Li_2025_ICCV}, all under the same protocol. We also considered PointSeg~\cite{he2025pointseg}
(training-free, but targeting 3D \emph{instance} segmentation under
detection-style mAP/AP metrics) and CLIP-FO3D~\cite{Zhang_2023_ICCV}
(trains a 3D encoder via distillation and does not, to our
knowledge, report ScanNet++ results). Neither is therefore included in Table~\ref{tab:scannetpp}. For the heritage dataset, no
prior annotation-free or training-free method reports results under a
comparable protocol. Instead we use a controlled internal comparison
(MiniLM vs.\ CLIP text encoders) as our primary evidence, detailed in
paragraph~\ref{para:semantic-truth}.

\textbf{Implementation.} We use Qwen2-VL~\cite{wang2024qwen2} for all ScanNet++ experiments.
For the heritage dataset, we use Gemini~\cite{team2023gemini}
instead, motivated by its substantially richer and more lexically
diverse descriptions on this domain (quantified in
Paragraph~\ref{para:vlmrichness}). We attribute this gap partly to
model scale: we use Qwen2-VL-2B, the smallest
model in its family, due to the compute constraints of a single
consumer GPU, whereas Gemini-2.5-Flash is substantially larger. This
asymmetry did not measurably affect ScanNet++, where Qwen2-VL alone
still performs competitively against 3D-trained baselines
(Table~\ref{tab:scannetpp}). Within each dataset, the VLM is held fixed
across the MiniLM and CLIP variants, so this choice does not affect
the fairness of that comparison. We report results using
all-MiniLM-L6-v2~\cite{wang2020minilm,reimers2019sentence} and, as an ablation, the substantially larger
CLIP ViT-L/14 text encoder~\cite{radford2021learning}.
Mean-inliers is our reported aggregation strategy throughout. All experiments run on a
single consumer laptop GPU (NVIDIA RTX~3070, 8\,GB VRAM; AMD Ryzen~9
5900HS; 40\,GB RAM), underscoring the pipeline's low hardware
requirements relative to methods that train a dedicated 3D encoder.

\begin{figure} [t]
\centering
\includegraphics[width=\linewidth]{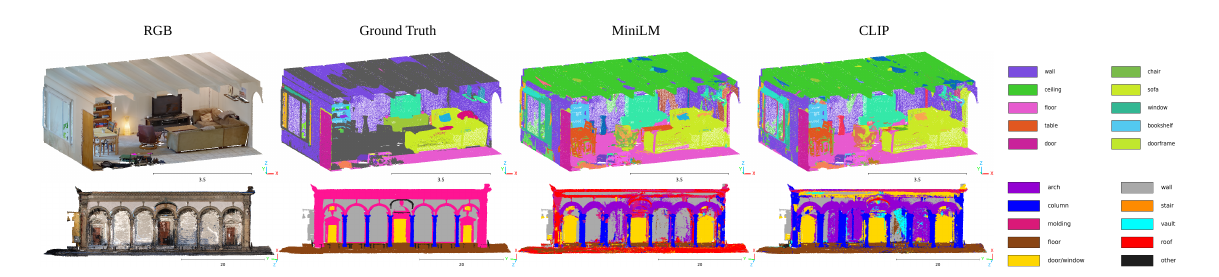}
\caption{Argmax semantic segmentation predictions, colored by class,
for MiniLM (left) and CLIP (right), on a ScanNet++ scene (top row,
100-class vocabulary) and a heritage building (bottom row, 10-class
ARCH vocabulary). Both variants require no 3D training on either
domain. Legends show ten classes per scene (10/100 for ScanNet++; all 10 for the heritage benchmark).}
  \label{fig:Qual_Results}
\end{figure}

\subsection{Quantitative \& Qualitative Results}
\label{subsec:results}
Table~\ref{tab:scannetpp} reports f-mIoU and f-mAcc
under the argmax protocol, with qualitative predictions shown in
Figure~\ref{fig:Qual_Results}. Our language-only embedding space
performs on par with the widely-used CLIP space, and in several
cases matches or exceeds methods trained on ScanNet, without any 3D
training of our own. MiniLM slightly
outperforms CLIP (16.83 vs.\ 16.17 f-mIoU) under otherwise identical
settings, isolating the effect of the text encoder alone. GoDeep also
attains substantially higher f-mAcc than every ScanNet-only baseline,
at the cost of increased false positives (Section~\ref{para:limitations}). As expected,
GoDeep falls behind Mosaic3D and SceneSplat when either is trained on
larger multi-dataset corpora or directly on ScanNet++, where training
on the target distribution provides a natural advantage. Table~\ref{tab:heritage} shows the reverse pattern: on the raw ARCH
vocabulary, CLIP outperforms MiniLM on all three metrics. Correcting a single mismatched prompt (\emph{floor}
$\to$~\emph{floor/grass}) dataset-wide, however, improves MiniLM far
more than CLIP across all three, reversing the mIoU ranking. A
second, independent correction (\emph{vault}~$\to$~\emph{vault/ceiling})
pushes MiniLM's mIoU to 35.96\%. Section~\ref{para:semantic-truth} traces this to a
systematic difference in how the two representations respond to
vocabulary quality.

\begin{table}[h]
\centering
\caption{ScanNet++ top-100 (argmax protocol). f-mIoU/f-mAcc exclude
wall, floor, ceiling. $^\dagger$Reproduced
by~\cite{Lee_2025_CVPR}, not reported in the original RegionPLC
paper.}
\label{tab:scannetpp}
\begin{tabular}{lccc}
\toprule
Method & 3D Training Data & f-mIoU & f-mAcc \\
\midrule
OpenScene~\cite{peng2023openscene} & ScanNet & 8.8 & 14.7 \\
RegionPLC~\cite{yang2024regionplc}$^\dagger$ & ScanNet & 11.3 & 20.1 \\
SceneSplat~\cite{Li_2025_ICCV} & ScanNet & 14.7 & 24.7 \\
Mosaic3D~\cite{Lee_2025_CVPR} & ScanNet & 16.2 & 27.1 \\
\textbf{GoDeep (MiniLM, ours)} & \textbf{-} & \textbf{16.83} & \textbf{39.08} \\
GoDeep (CLIP, ours) & - & 16.17 & 39.06 \\
\midrule
Mosaic3D~\cite{Lee_2025_CVPR} & multi-dataset (5.6M) & 18.0 & 29.0 \\
SceneSplat~\cite{Li_2025_ICCV} & ScanNet++ & 26.8 & 45.3 \\
SceneSplat~\cite{Li_2025_ICCV} & SN+SN+++MP3D & \textbf{28.4} & \textbf{50.0} \\
\bottomrule
\end{tabular}
\end{table}

\begin{table}[h]
\centering
\footnotesize
\setlength{\tabcolsep}{4pt}
\caption{Cultural heritage benchmark, 5 buildings~\cite{pellis2025photogrammetric},
10 ARCH classes~\cite{matrone2020benchmark}, including \emph{other}.
\emph{Corrected} replaces \emph{floor}~$\to$~\emph{floor/grass}
dataset-wide; \emph{Both} additionally replaces
\emph{vault}~$\to$~\emph{vault/ceiling}. $\Delta$ shown relative to
raw.}
\label{tab:heritage}
\resizebox{\linewidth}{!}{%
\begin{tabular}{lcccccc}
\toprule
& \multicolumn{2}{c}{mIoU} & \multicolumn{2}{c}{mAcc} & \multicolumn{2}{c}{mPrec} \\
\cmidrule(lr){2-3} \cmidrule(lr){4-5} \cmidrule(lr){6-7}
Variant & raw & corr.\ ($\Delta$) & raw & corr.\ ($\Delta$) & raw & corr.\ ($\Delta$) \\
\midrule
MiniLM & 22.70 & 32.98 (+10.28) & 49.83 & 54.42 (+4.59) & 40.48 & \textbf{49.80 (+9.32)} \\
MiniLM (Both) & 22.70 & \textbf{35.96 (+13.26)} & 49.83 & \textbf{56.31 (+6.48)} & 40.48 & 49.27 (+8.79) \\
CLIP & 25.92 & 28.48 (+2.56) & 50.55 & 52.90 (+2.35) & 46.33 & 48.42 (+2.09) \\
CLIP (Both) & 25.92 & 29.47 (+3.55) & 50.55 & 54.01 (+3.46) & 46.33 & 49.46 (+3.13) \\
\bottomrule
\end{tabular}
}
\end{table}

\section{Ablation Studies}
\label{sec:ablation}

We organize this section around three questions. Firstly, how
faithfully do language space embeddings track physical content
compared to CLIP based alternatives, both under controlled vocabulary
corrections (Paragraph~\ref{para:semantic-truth}) and on genuinely
out-of-vocabulary objects (Paragraph~\ref{para:ood})? Secondly, what
design choices drive this behavior: the choice of VLM
(Paragraph~\ref{para:vlmrichness}, \ref{para:vlmconsistency}) and the
choice of aggregation strategy (Paragraph~\ref{par:aggcompar})? Thirdly,
how transparent is the resulting representation
(Paragraph~\ref{para:explainability})?

\ablationhead{Semantic Truth vs. Annotation Convention}\label{para:semantic-truth}
Table~\ref{tab:heritage} shows that correcting a single systematic
vocabulary mismatch (\emph{floor}~$\to$~\emph{floor/grass}), applied
dataset-wide, improves MiniLM roughly $4\times$ more than CLIP in
overall mIoU (+10.28 vs.\ +2.56). To verify this is not an isolated
effect, we apply a second, independent correction
(\emph{vault}~$\to$~\emph{vault/ceiling}) to a different class
entirely. Table~\ref{tab:semantictruth} shows that the same asymmetry
holds: MiniLM's overall mIoU gain from this second correction alone
(+2.93) is again roughly $3\times$ CLIP's (+0.89), and applying both
corrections jointly is close to additive for MiniLM, but remains
small for CLIP throughout (+3.55).

Each correction produces a distinct cascading effect, largely
confined to its own semantically related class: correcting
\emph{floor} raises MiniLM's \emph{column} IoU by +29.00 while
leaving \emph{vault/ceiling} untouched (+0.00), whereas correcting
\emph{vault} raises \emph{arch} by +5.56 while leaving \emph{column}
untouched (+0.03). This class-specific, non-overlapping pattern is
itself evidence that the effect reflects genuine semantic structure
rather than noise. We trace the mechanism behind the larger of the
two, the \emph{column} gain: 3.04M ground-truth \emph{floor/grass}
points were incorrectly predicted as \emph{column} by MiniLM before
correction, because the uncorrected \emph{floor} prompt was a poor
semantic match for these points, pushing them toward whatever prompt
was next closest. Correcting the prompt reclaims 96.3\% of these
points for \emph{floor/grass}, which directly reduces
\emph{column}'s false positives and raises its IoU. The same leak
affects CLIP (690K points), but correction reclaims essentially none
of it (the leak grows slightly, by 1.7\%), consistent with CLIP's
comparatively muted response to vocabulary correction throughout
Table~\ref{tab:semantictruth}. Together, these two independent
corrections indicate that MiniLM's language-space embeddings track
physical content more precisely than CLIP's, a property that becomes
a liability only when the evaluation vocabulary itself is imprecise
(Contribution~3).

\begin{table}[h]
\centering
\caption{Effect of two independent vocabulary corrections
(\emph{floor}~$\to$~\emph{floor/grass}, \emph{vault}~$\to$~\emph{vault/ceiling}),
applied separately and jointly, across the full 5-building heritage
dataset. $\Delta$IoU shown per class relative to the uncorrected
baseline.}
\label{tab:semantictruth}
\begin{tabular}{lcccccc}
\toprule
& \multicolumn{3}{c}{MiniLM $\Delta$IoU} & \multicolumn{3}{c}{CLIP $\Delta$IoU} \\
\cmidrule(lr){2-4} \cmidrule(lr){5-7}
Class & grass & ceiling & Both & grass & ceiling & Both \\
\midrule
floor/grass   & +45.32 & +0.12 & +45.63 & +22.27 & $-$0.04 & +22.03 \\
vault/ceiling & +0.00  & +26.15 & +26.24 & $-$0.67 & +9.94 & +9.84 \\
column        & +29.00 & +0.03 & +29.11 & $-$0.10 & $-$0.29 & $-$0.43 \\
stair         & +15.02 & +0.00 & +15.05 & +0.92 & $-$0.57 & +0.06 \\
roof          & +8.21  & $-$0.85 & +7.38 & +0.03 & $-$1.54 & $-$1.52 \\
arch          & +0.01  & +5.56 & +5.64 & $-$0.02 & +0.66 & +0.57 \\
\midrule
Overall mIoU (10 classes) & +10.28 & +2.93 & \textbf{+13.26} & +2.56 & +0.89 & +3.55 \\
\bottomrule
\end{tabular}
\end{table}

\ablationhead{Out-of-Distribution Evaluation}\label{para:ood} We test whether GoDeep
recognizes objects outside the 100-class ScanNet++ vocabulary, using
25 out-of-vocabulary classes spanning structural, technical,
furniture, and fixture categories, under three conditions: present
(the object's ground-truth points exist in the scene), absent (a
different scene where they do not), and extreme (25 categories
irrelevant to any indoor scene, as a calibration floor).

Both encoders correctly separate present from absent from extreme on
average (Table~\ref{tab:oodcalibration}), but MiniLM's present/absent
margin is $2.4\times$ CLIP's ($+0.143$ vs.\ $+0.061$), and CLIP's
absent scores are barely distinguishable from its extreme baseline
($+0.008$ vs.\ MiniLM's $+0.092$), indicating weaker calibration
between ``plausible but absent'' and ``nonsensical'' queries. This
separation is not universal: 4/25 classes invert (absent scoring
marginally higher than present) for each encoder, with \emph{fake
ceiling} inverting for both, suggesting a possible confound in that
particular absent-scene assignment.

We further compare localization once an object is present, via
per-scene, per-encoder adaptive threshold sweeps (50th--90th
percentile) and binary IoU against ground truth. MiniLM wins more
class-threshold comparisons (72/125 vs.\ 53/125) and achieves higher
overall mean IoU (0.204 vs.\ 0.189), though the advantage that it is
category-dependent: structural and fixture objects favor MiniLM
consistently (6/6 classes), furniture shows no meaningful difference,
and CLIP's largest wins occur on two objects with generic component
names (\emph{cable tray}, \emph{folding screen}). This mirrors
paragraph~\ref{para:semantic-truth}: MiniLM's language-space embeddings
respond more strongly to vocabulary precision, benefitting more when
it is high and, by the same mechanism, suffering more when a query is
inherently ambiguous, while CLIP's more diffuse joint embedding space
is comparatively insensitive to wording precision in either
direction.

\begin{table}[h]
\centering
\caption{Out-of-distribution calibration and localization on
ScanNet++, 25 classes per suite. \emph{Present/Absent/Extreme}: mean
similarity of each scene's top-100 highest-scoring points (a
noise-robust proxy for confidence), for an object confirmed present,
confirmed absent from that scene, and entirely unrelated to any
indoor scene, respectively. $\Delta_{P-A}$: present$-$absent gap
(higher indicates sharper discrimination between "object here" and
"object elsewhere"). \emph{Loc.\ win-rate}: fraction of 125
class$\times$threshold comparisons (25 classes $\times$ 5
percentile thresholds, 50th--90th) where an encoder achieves higher
binary IoU against ground truth. \emph{Mean IoU}: overall mean IoU
across all comparisons.}
\label{tab:oodcalibration}
\begin{tabular}{lcccccc}
\toprule
Encoder & Present & Absent & Extreme & $\Delta_{P-A}$ & Loc.\ win-rate & Mean IoU \\
\midrule
MiniLM & \textbf{0.704} & 0.561 & 0.469 & \textbf{+0.143} & \textbf{72/125} & \textbf{0.204} \\
CLIP   & 0.590 & 0.529 & 0.522 & +0.061 & 53/125 & 0.189 \\
\bottomrule
\end{tabular}
\end{table}

\ablationhead{VLM Description Richness}\label{para:vlmrichness} We compare Qwen2-VL-2B and Gemini-2.5-Flash on the same 748 images
(Table~\ref{tab:vlmrichness}), reporting MTLD~\cite{mccarthy2010mtld},
a length-robust lexical diversity measure computed as the mean
number of words required for the running type-token ratio to drop to
a fixed threshold (0.72), averaged over forward and backward passes
through the text.
Gemini produces descriptions that are $4\times$ longer on average and
more lexically diverse by every measure we compute: MTLD is
$2.1\times$ higher, and Gemini uses $3.4\times$ more unique
adjectives and $3.1\times$ more unique nouns, both directly relevant
to the material, color, and part-level attributes our pipeline
extracts. This gap is expected given model scale: Qwen2-VL-2B is the
smallest model in its family, chosen to fit within a single consumer
GPU's compute budget, whereas Gemini-2.5-Flash is substantially
larger. The asymmetry did not measurably affect ScanNet++, where
Qwen2-VL alone remains competitive against 3D-trained baselines
(Table~\ref{tab:scannetpp}), suggesting that model scale matters more
for domains with unusually specialized vocabulary (e.g., architectural
terminology) than for general indoor scenes. A larger Qwen2-VL variant (7B/72B) would likely narrow this gap,
though at a compute cost beyond our single GPU, however we leave this to
future work.

\begin{table}[h]
\centering
\caption{Lexical richness of VLM descriptions, same 748 images
(1\_SC scene). MTLD is a length-robust lexical diversity measure;
unique adjectives/nouns are counted by lemma.}
\label{tab:vlmrichness}
\begin{tabular}{lccccc}
\toprule
VLM & Words/desc.\ (mean) & MTLD & Unique words & Unique adj. & Unique nouns \\
\midrule
Qwen2-VL-2B & 231.0 & 46.81 & 1,968 & 397 & 785 \\
Gemini-2.5-Flash & 914.4 & 97.41 & 6,035 & 1,347 & 2,408 \\
\bottomrule
\end{tabular}
\end{table}

\ablationhead{VLM Description Consistency}\label{para:vlmconsistency} We compare descriptions pairwise across runs using two measures:
exact-match rate, the fraction of image descriptions that are
byte-identical between two runs, and Jaccard similarity, the ratio
of shared to total unique words between two descriptions of the same
image (1.0 = identical vocabulary, 0.0 = no overlap). Although generation nominally
uses sampling (temperature $=0.7$), we observe that Qwen2-VL-2B,
loaded in 4-bit quantization, produces byte-identical descriptions
across four independent runs on the same 360 images (100\% exact
match, mean pairwise Jaccard similarity of $1.000$), with no random
seed fixed in our implementation. We attribute this to quantization
narrowing the output probability distribution sufficiently that
sampling collapses to argmax selection in practice, though we did
not verify this mechanism directly. Regardless of cause, the
practical implication is that our language-space representation is
fully reproducible under our deployment configuration.

\begin{table}[h]
\centering
\caption{Consistency of Qwen2-VL-2B descriptions across 4 independent
runs on the same 360 ScanNet++ images, despite nominal sampling
(temperature $=0.7$). Per run lexical statistics are pooled across
runs; cross-run metrics compare descriptions pairwise.}
\label{tab:vlmconsistence}
\begin{tabular}{lc}
\toprule
Metric & Qwen2-VL-2B \\
\midrule
Images Evaluated & 360 \\
Independent Runs & 4 \\
Words / Description (mean) & 272.5 \\
MTLD (Lexical Diversity) & 32.6 \\
Unique Words & 1,292 \\
\midrule
Mean Pairwise Jaccard Similarity & 1.000 \\
Exact-Match Rate (byte-identical) & 100.0\% \\
\bottomrule
\end{tabular}
\end{table}

\begin{figure}[t]
\centering
\includegraphics[width=\linewidth]{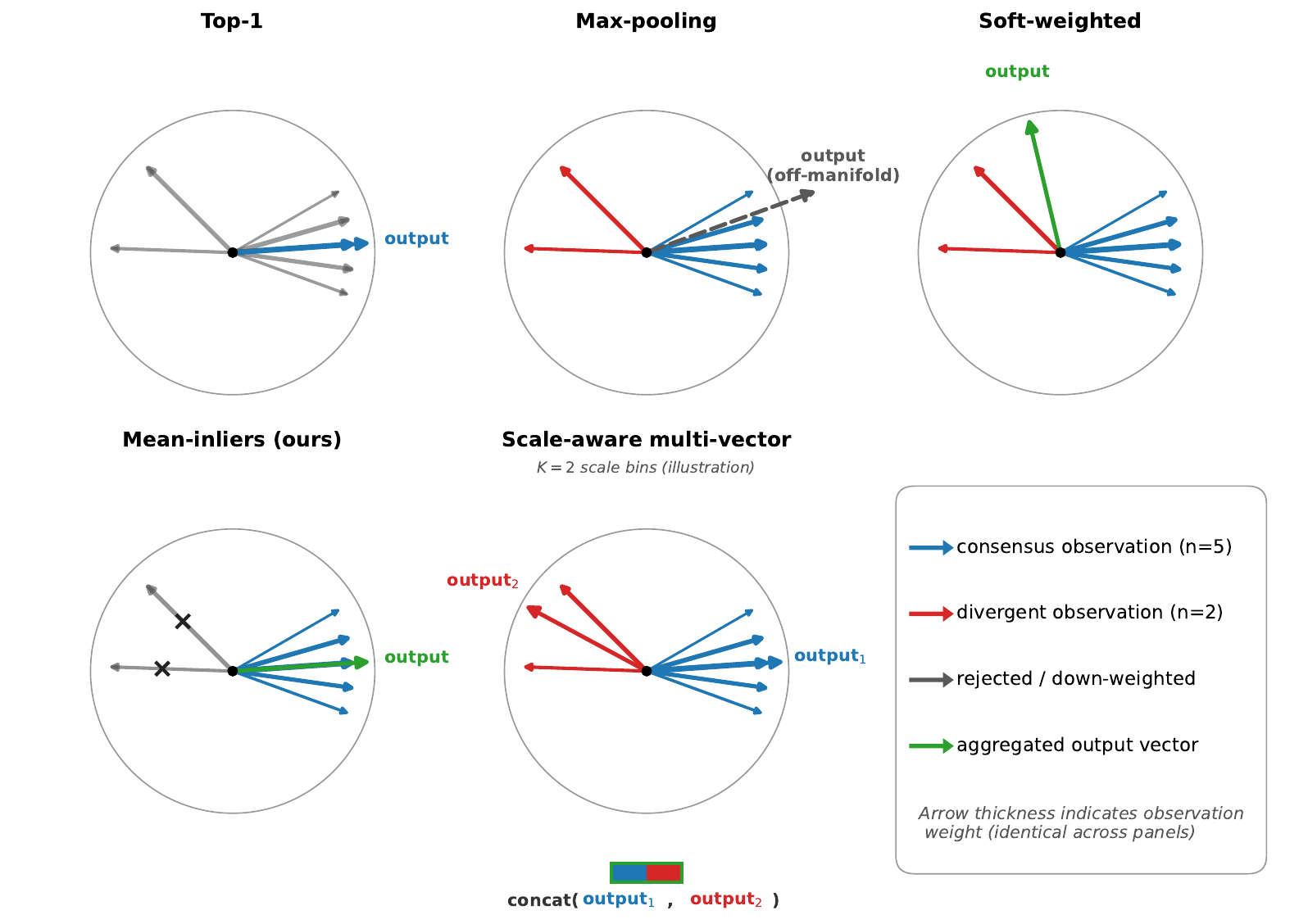}
\caption{Schematic comparison of the five aggregation strategies
(Paragraph~\ref{par:aggcompar}) on an identical set of seven
observations: five angularly clustered ``consensus'' observations and
two angularly divergent ones, with arrow thickness indicating
observation weight $w_i$ (identical across panels; illustrative
weights, not drawn from actual pipeline data). \emph{Top-1} keeps
only the highest-weight observation. \emph{Max-pooling} ignores
weights entirely and produces a vector off the unit-norm embedding
manifold (dashed). \emph{Soft-weighted} amplifies the divergent
observations rather than suppressing them. \emph{Mean-inliers} (ours)
rejects them as outliers. \emph{Scale-aware multi-vector}, shown here
with $K{=}2$ scale bins for illustration ($K{=}3$ in our reported
experiments), preserves both groups as separate output vectors that
are concatenated rather than merged.}
\label{fig:aggregation_circles}
\end{figure}

\ablationhead{Aggregation Strategies and Comparison}\label{par:aggcompar} Beyond mean-inliers (Section~\ref{subsec:mv3dlift}),
we evaluate four alternative aggregation strategies for combining a
point's weighted set of natural-language observations
$\{(v_i, w_i)\}$ into a single embedding (Figure~\ref{fig:aggregation_circles}). \emph{Top-1} selects only
the single observation with the highest weight $w_i$, discarding all
others: the simplest possible choice, but with no robustness to a
single mis-grounded or hallucinated observation. \emph{Max-pooling}
takes the element-wise maximum across all observed vectors,
independent of their weights entirely. This is common in
convolutional feature aggregation but disrupts the cosine-similarity
geometry of a sentence-embedding space, since the resulting vector no
longer corresponds to any single coherent point on the embedding
manifold. \emph{Soft-weighted} aggregation takes the opposite
philosophy from mean-inliers: rather than rejecting observations that
diverge from the consensus, it computes an initial weighted mean,
then boosts the weight of each observation in proportion to how much
it diverges from that mean, amplifying rather than suppressing
minority evidence. \emph{Scale-aware multi-vector} aggregation groups
a point's observations by how frequently each underlying phrase
occurs across the entire scene (independent of that specific point),
into $K{=}3$ bins from most to least common; mean-inliers pooling is
then applied independently within each bin, and the $K$ resulting
vectors are concatenated. This prevents a rare but precise
observation (e.g., ``ornate capital'') from being diluted by a
frequent, generic one (e.g., ``column''), the same mechanism
underlying the rank-shift evidence in Section~\ref{para:explainability}.

We compare all five strategies on the same scene (1\_SC), MiniLM held
fixed, RAW embeddings throughout (Table~\ref{tab:aggregation}).
Scale-aware multi-vector achieves the highest mIoU (30.18), exceeding
mean-inliers by $+4.36$ points, consistent with the explainability
mechanism just described. Soft-weighted also outperforms mean-inliers,
suggesting that in a heterogeneous heritage scene, observations that
mean-inliers treats as outliers are often valid, rare evidence rather
than noise. Max-pooling performs worst, consistent with it ignoring
the weighting scheme entirely and disrupting the embedding space's
geometry. These results are based on a single scene and come with a
substantial storage cost for scale-aware aggregation (24.3\,GB vs.\
6.5--8\,GB for the other strategies, from the $K{=}3$ concatenated
representation). We therefore report mean-inliers as our primary
method throughout this paper, while these findings motivate
full-dataset validation of scale-aware aggregation as future work.

\begin{table}[h]
\centering
\caption{Aggregation strategy comparison, scene 1\_SC, MiniLM fixed,
RAW embeddings. Storage measured for the full scene's embedding file.}
\label{tab:aggregation}
\begin{tabular}{lcccc}
\toprule
Method & mIoU & mAcc & mPrec & Storage \\
\midrule
\textbf{Scale-aware multi-vector} & \textbf{30.18} & \textbf{57.37} & \textbf{47.65} & 24.3\,GB \\
Soft-weighted & 27.27 & 52.92 & 47.39 & $\sim$7\,GB \\
Mean-inliers (reported) & 25.82 & 51.28 & 41.23 & $\sim$6.8\,GB \\
Top-1 & 21.67 & 47.06 & 32.32 & $\sim$8\,GB \\
Max-pooling & 20.68 & 41.80 & 46.40 & $\sim$6.5\,GB \\
\bottomrule
\end{tabular}
\end{table}

\ablationhead{Explainability of the Weighting Scheme}\label{para:explainability} Because every intermediate representation remains discrete text
(Section~\ref{sec:method}), a prediction can be traced end to end: from the VLM
description of a view, to the text assigned to each mask, to the
weighted text observed at each point, to the final sentence
embedding it produces. This lets us inspect not only how a point was labeled, but why. We isolate all points
belonging to column capitals in one heritage scene (1\_SC) and rank their
associated descriptions either by raw observation frequency or by
the weight from Eq.~\eqref{eq:weight}. Descriptions naming the
capital explicitly rank low by frequency (\emph{``Ornate
Capital''}, rank 14; \emph{``Ornate Classical Capital''}, rank 17)
but rise sharply once weighted (rank 4 and 6, respectively), while
generic column descriptions with comparable frequency drop
correspondingly (e.g., \emph{``Columns''}, rank 3~$\to$~8). This weighting is therefore a substantive design choice: it
systematically prioritizes precise, well-observed evidence over
evidence that is merely frequent (Contribution~4).

\section{Discussion}
\label{sec:discuss}

GoDeep decouples 3D scene understanding from CLIP's joint
vision-language space, lifting VLM descriptions directly into a
sentence-embedding space instead (Contributions~1--2), matching
strong annotation-free ScanNet++ baselines without 3D training. No
comparable baseline exists for our heritage benchmark, where evidence
instead comes from a controlled internal comparison.

That comparison grounds Contribution~3: two independent
vocabulary corrections, dataset-wide, both show MiniLM responding
far more strongly than CLIP (Paragraph~\ref{para:semantic-truth}), with
a traced mechanism rather than a bare correlation. The
out-of-distribution evaluation extends this to genuinely unseen
vocabulary (Paragraph~\ref{para:ood}): MiniLM discriminates presence
from absence more sharply, though localization is category-dependent
rather than uniformly favoring either encoder.

Contribution~4 rests on every intermediate representation
remaining discrete text, making predictions traceable to specific
evidence (Paragraph~\ref{para:explainability}): this traceability is
actionable, not only diagnostic, since identifying that mean-inliers
discards rare evidence directly motivated scale-aware multi-vector
aggregation (Paragraph~\ref{par:aggcompar}).

\paragraph{Limitations:}\label{para:limitations} Querying a VLM and LLM
per image adds computational cost ($\sim$21 sec and $\sim$51 sec per
ScanNet++ image, respectively, on a single consumer GPU), though far
less than curating 3D training corpora. Excluding this one-time
step, projecting and aggregating text features takes on average
$\sim$14.5 minutes per ScanNet++ scene. GoDeep's higher f-mAcc on
ScanNet++ (Table~\ref{tab:scannetpp}) reflects a bias toward recall
over precision, consistent with argmax classification over dense,
unconstrained VLM descriptions. Gemini, used for the heritage VLM,
is closed-source and less reproducible than Qwen2-VL, and the
heritage results rely on an internal rather than external
comparison, since no prior method reports results under a comparable
protocol on this domain. Our representations are memory-intensive in
general: even mean-inliers, our reported strategy, requires
$\sim$6.8\,GB for a scene of $\sim$5M points. Scale-aware multi-vector
aggregation (Paragraph~\ref{par:aggcompar}) is the most extreme case,
at 24.3\,GB on the same scene ($3.5\times$ mean-inliers, from its
$K{=}3$ concatenated representation). Aggregation comparison results
are also based on a single scene, and we leave full-dataset
validation, along with reducing this footprint, to future work.

\section{Conclusion}\label{sec:conclusion}
We presented GoDeep, an annotation-free 3D scene understanding
pipeline that lifts structured VLM descriptions into a pure
sentence-embedding space, requiring no 3D training corpus or
dedicated 3D encoder. On ScanNet++, GoDeep matches strong
annotation-free baselines trained on ScanNet. On a 5-building
cultural heritage benchmark, two independent vocabulary corrections
show that language-space embeddings track physical content more
faithfully than CLIP-based embeddings, reversing the raw benchmark
ranking. This fidelity extends to genuinely out-of-vocabulary
objects and generalizes across five aggregation strategies, one of
which (scale-aware multi-vector) exceeds our reported configuration
at a storage cost we leave to future work to reduce. Throughout,
GoDeep's predictions remain traceable to the specific natural-language
evidence that produced them, a property we use not only to explain
individual predictions but to diagnose and improve the pipeline
itself.

\section*{Acknowledgements}
The publication of this paper is supported by the Research Project "Photogrammetry and
Geoinformatics" no. 95030400 from the National Technical University of
Athens. The authors would like to thank Dr. Katerina Adam and the anonymous reviewers for their constructive comments.

%
%
\bibliographystyle{splncs04}
\bibliography{main}
\end{document}